\documentclass[conference]{IEEEtran}
\IEEEoverridecommandlockouts
\usepackage{cite}
\usepackage{amsmath,amssymb,amsfonts}
\usepackage{algorithmic}
\usepackage{graphicx}
\usepackage{textcomp}
\usepackage{xcolor}
\usepackage{array}
\usepackage{booktabs}
\usepackage{multirow}
\def\BibTeX{{\rm B\kern-.05em{\sc i\kern-.025em b}\kern-.08em
    T\kern-.1667em\lower.7ex\hbox{E}\kern-.125emX}}
\begin{document}

\title{SafeCtrl: Region-Aware Safety Control for Text-to-Image Diffusion via Detect-Then-Suppress}


\author{
    \IEEEauthorblockN{Lingyun Zhang$^{1,2,*}$,
    Yu Xie$^{3,*}$,
    Zhongli Fang$^{1,2}$,
    Yu Liu$^{1,2}$,
    Ping Chen$^{2,\dagger}$ ,
    \thanks{$^*$ Equal contribution. $^\dagger$ Corresponding author.
    }}
    \IEEEauthorblockA{
    \textsuperscript{1}School of Computer Science and Technology, Fudan University, China 
  \textsuperscript{2}Institute of Big Data, Fudan University,  China \\
  \textsuperscript{3} yxie93@nuist.edu.cn, Nanjing University of Information Science and Technology,  China
    }
}

\maketitle

\begin{abstract}
The widespread deployment of text-to-image diffusion models is significantly challenged by the generation of visually harmful content, such as sexually explicit content, violence, and horror imagery. Common safety interventions, ranging from input filtering to model concept erasure, often suffer from two critical limitations: (1) a severe trade-off between safety and context preservation, where removing unsafe concepts degrades the fidelity of the safe content, and (2) vulnerability to adversarial attacks, where safety mechanisms are easily bypassed.
To address these challenges, we propose SafeCtrl, a Region-Aware safety control framework operating on a Detect-Then-Suppress paradigm.
Unlike global safety interventions, SafeCtrl first employs an attention-guided Detect module to precisely localize specific risk regions.
Subsequently, a localized Suppress module, optimized via image-level Direct Preference Optimization (DPO), neutralizes harmful semantics only within the detected areas, effectively transforming unsafe objects into safe alternatives while leaving the surrounding context intact.
Extensive experiments across multiple risk categories demonstrate that SafeCtrl achieves a superior trade-off between safety and fidelity compared to state-of-the-art methods.
Crucially, our approach exhibits improved resilience against adversarial prompt attacks, offering a precise and robust solution for responsible generation.
\end{abstract}


\section{Introduction}
\label{sec:introduction}

Diffusion models~\cite{ ho2020denoising,rombach2022high} have revolutionized visual synthesis but pose significant risks by generating harmful content~\cite{birhane2021multimodal,qu2023unsafe}. The challenge is particularly acute for {visually harmful risks}, such as nudity, violence, and horror imagery, which are readily learned from web-scale data~\cite{schuhmann2022laionb}. As these models are deployed, ensuring safety without compromising their core generative capabilities remains a critical barrier.

Concept erasure methods, such as Fine-tuning~\cite{gandikota2023erasing, li2025responsible} or Inference Guidance~\cite{schramowski2023safe}, aim to suppress unsafe concepts in diffusion models across the entire generated image. However, they suffer from two fundamental limitations: (1) {Context Degradation}: Removing a concept globally often alters the safe background or unrelated objects, leading to a poor trade-off between safety and fidelity. (2) {Vulnerability}: These methods heavily rely on text conditioning. If adversarial prompts hide the harmful intent, the safety mechanism fails to trigger.
More recently, Concept Replacer (CR)~\cite{zhang2025concept} proposed a localized pipeline. While promising, CR necessitates a duplicated U-Net for localization and relies on ``hard replacement'', which often introduces semantic artifacts and rigidity.

\begin{figure}[t]
    \centering
    \includegraphics[width=0.9\linewidth]{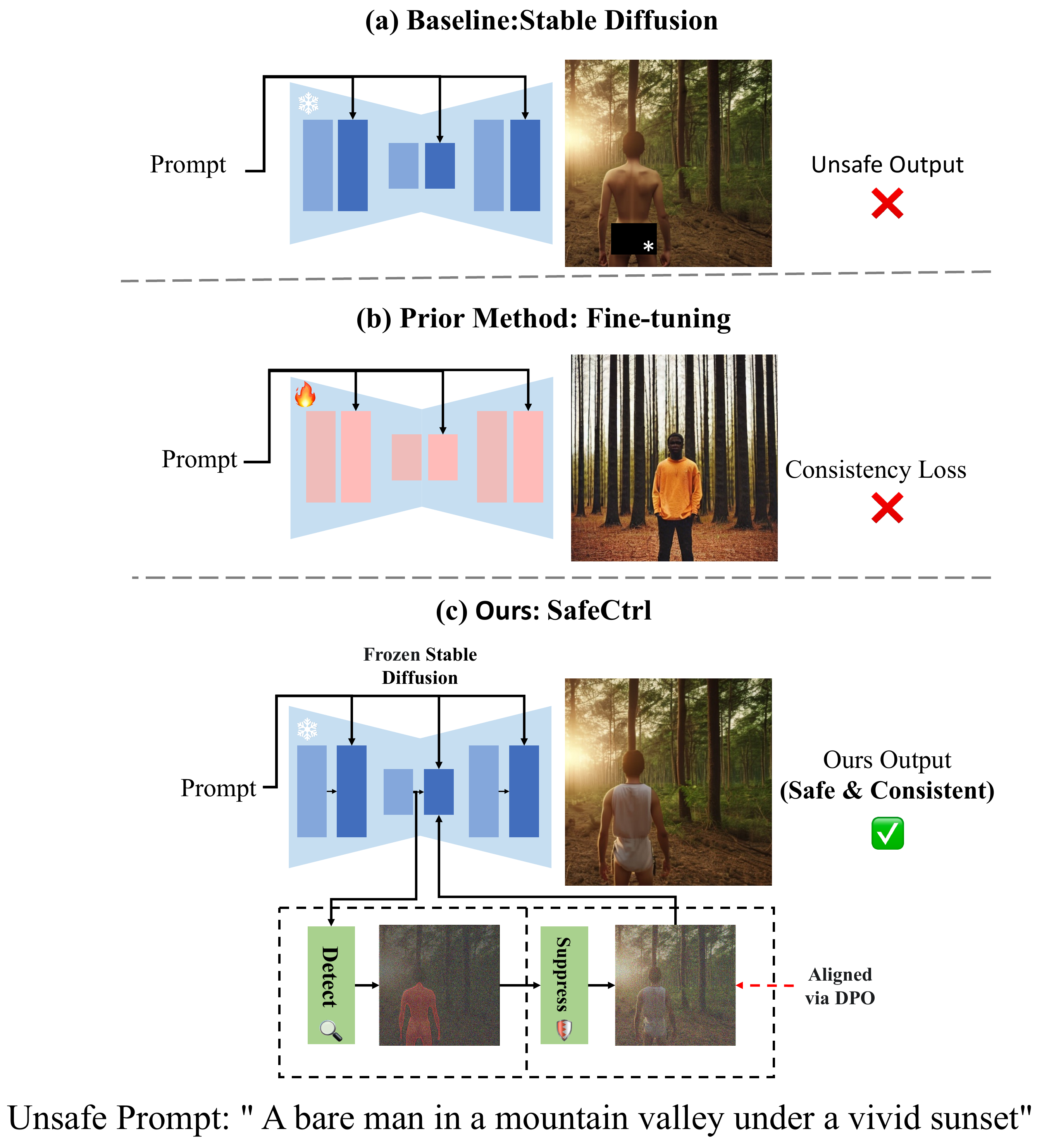} 
    \caption{
        \textbf{Solving the safety and context preservation trade-off.} 
        (a) The original {Stable Diffusion} generates unsafe content (e.g., sexually explicit content) from a harmful prompt. 
        (b) {Prior Global Fine-tuning} methods (e.g., ESD) remove the concept but suffer from severe {context degradation}, altering the background, lighting, and identity. 
        (c) {SafeCtrl (Ours)} employs a {Region-Aware} ``Detect-Then-Suppress'' paradigm. By keeping the base model frozen (Blue blocks) and using a external module (Green blocks) trained via DPO, SafeCtrl precisely localizes and neutralizes risk, ensuring safety while strictly preserving the original context and artistic intent.
    }
    \label{fig:teaser}
    \vspace{-4mm} 
\end{figure}

\begin{figure*}[t!]
    \centering
    \includegraphics[width=1.\textwidth]{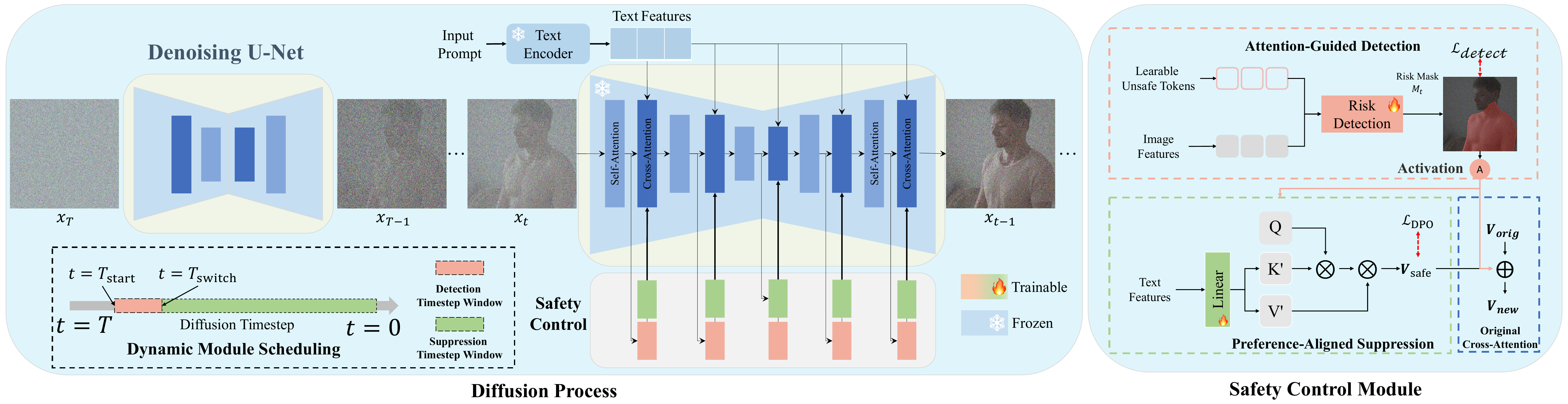}
    \caption{
        \textbf{The overall architecture of the SafeCtrl framework.} 
        \textbf{(Left)} SafeCtrl is instantiated as a set of external modules that operate in parallel with frozen U-Net.
        The safety control is adaptively activated within specific diffusion stages: the {Detection Timestep Window} $[T_{\text{start}}, T_{\text{switch}} ]$ and the \textit{Suppression Timestep Window} $[T_{\text{switch}}, 0 ]$. 
        \textbf{(Right)} A detailed view inside a single Safety Control Module, illustrating the {Detect-Then-Suppress} paradigm. The {Risk Detection} component first analyzes image features to compute risk mask.The suppression process is governed by an {activation trigger} ($A$). Only if a risk is detected does the Preference-Aligned Suppression component activate, computing a safety-guided attention output (\(V_{safe}\)) to the original cross-attention layer.
    }
    \label{fig:method_architecture}
\end{figure*}

To overcome these limitations, we introduce a method called SafeCtrl.
Our core insight is that for risks like explicit content or horror imagery,  the diffusion model's internal features reveal the harmful regions even when textual prompts are adversarial. 
Unlike global safety methods, SafeCtrl first employs an attention-guided {Detect} module to precisely localize unsafe regions. Subsequently, a {Suppress} module performs a localized intervention to neutralize harmful semantics {only within the detected mask}.
To ensure coherent generation, we introduce a training strategy using image-level Direct Preference Optimization (DPO), allowing the module to learn nuanced suppression behaviors without costly pixel-level annotations.
This region-aware design offers several advantages. It preserves the integrity of the safe context and anchors defense on visual features rather than just text.
Our main contributions are:
\begin{itemize}
    \item We propose SafeCtrl, a {Region-Aware} framework that decouples safety into a `Detect-Then-Suppress' process.
    \item We introduce an attention-guided detection mechanism that localizes unsafe concepts, providing {greater robustness to adversarial prompt attacks} than existing baselines.
    \item We devise a DPO-based training strategy that achieves precise local suppression using only image-level supervision. Experiments show SafeCtrl achieves a {superior trade-off} between safety and context preservation.
\end{itemize}

\section{Related Work}
\label{sec:related_work}

\textbf{Global Safety Interventions.}
The dominant paradigm for safety involves suppressing harmful concepts across the entire image, either by fine-tuning weights or guiding inference. 
Inference-time methods like SLD~\cite{schramowski2023safe} manipulate unconditional score estimates to steer generation away from unsafe directions.
Fine-tuning approaches, such as ESD~\cite{gandikota2023erasing}, UCE~\cite{gandikota2024unified}, and MACE~\cite{lu2024mace}, modify U-Net weights (or LoRA~\cite{hu2022lora} modules) to ``forget'' specific concepts. Similarly, RDM~\cite{li2025responsible} learns a safe semantic vector to shift the text embedding globally.
However, these global interventions suffer from two critical limitations. 
First, they incur {context degradation}. Applying a static, global constraint often alters safe background elements, leading to a poor trade-off between safety and fidelity. 
Second, they exhibit {vulnerability to adversarial attacks}. Since they heavily rely on textual concept matching, they can be easily bypassed by adversarial prompts or obfuscated tokens.

\textbf{Region-Aware Control.}
To enable more precise interventions, recent works leverage cross-attention maps for localized control~\cite{couairon2022diffedit,mou2024t2i,ye2023ip}. 
In the safety domain, Concept Replacer (CR)~\cite{zhang2025concept} pioneers a Localize-and-Replace pipeline. 
Despite its decoupling design, CR has fundamental flaws. 
First, it relies on a ``hard replacement'' paradigm (e.g., requiring a specific safe target word), which often introduces semantic artifacts and lacks the flexibility to generate contextually natural alternatives. 
Second, CR necessitates a duplicated, heavyweight U-Net for localization, doubling the inference memory overhead.
In contrast, {SafeCtrl} reuses internal attention features to eliminate the need for an auxiliary U-Net and employs DPO-based suppression to learn natural, context-aware safety transformations without explicit replacement prompts.

\section{Method}

\begin{figure*}[t!]
    \centering
    \includegraphics[width=0.9\textwidth]{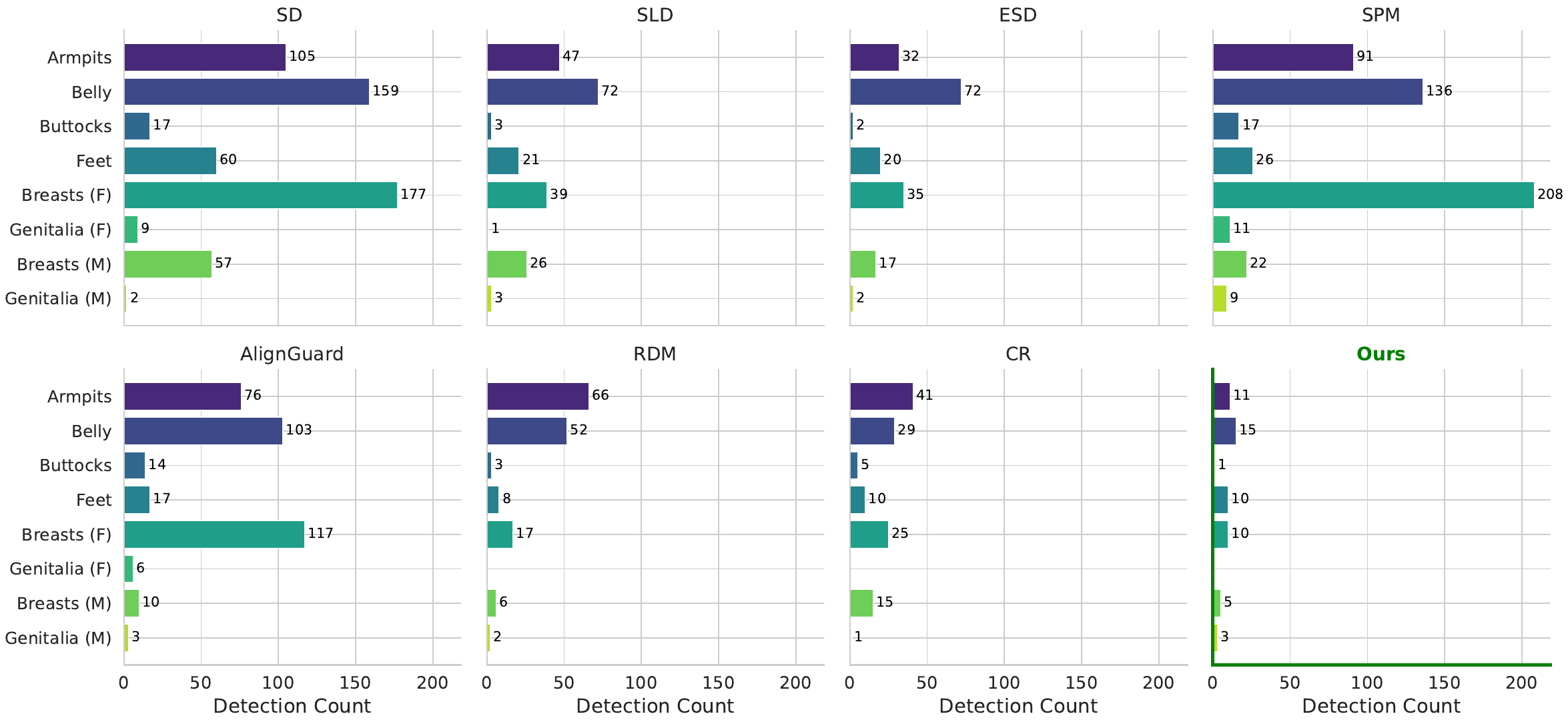}
    \caption{
        \textbf{Detailed comparison of NudeNet detection counts on the I2P benchmark.}
        Each subplot corresponds to a different safety method, showing the number of generated images flagged for various nudity-related categories (lower is better). 
    }
    \label{fig:app_nudenet_grid_2x4}
\end{figure*}

\subsection{Framework Overview}
We propose {SafeCtrl}, a region-aware safety framework operating on a ``Detect-Then-Suppress'' paradigm as illustrated in Figure~\ref{fig:method_architecture}. 
Instead of modifying the pre-trained U-Net weights, SafeCtrl introduces two lightweight modules: a {Detect Module} ($\mathcal{D}$) and a {Suppress Module} ($\mathcal{S}$).
Crucially, we employ a {Dynamic Scheduling} strategy to minimize inference overhead.
Diffusion models typically establish semantic structure in the early denoising stages \cite{hertz2022prompt}.
Therefore, our {Detect Module} is only activated during a short window to accumulate a stable spatial risk mask $M$.
Once a risk is confirmed, the {Suppress Module} activates for the subsequent steps, fusing safe features only within the detected regions.

\subsection{Attention-Guided Detection}
The core of our robustness lies in anchoring detection on the model's internal visual features. We implement the Detect module as an attention-based detection.

We extract features directly from the frozen U-Net's intermediate layers.
Let $f \in \mathbb{R}^{H \times W \times C}$ be the spatial features. 
The cross-attention map $A_{\text{cross}}$, which aligns visual regions with textual concepts, is derived from the query $Q = W_Q f$ and text key $K = W_K c$: 
\begin{equation}
    A_{\text{cross}} = \text{Softmax}(Q K^T / \sqrt{d}).
\end{equation}
where $W_Q$ and $W_K$ are projection matrices, c is the prompt embedding.
Concurrently, the self-attention map $A_{\text{self}}$, capturing object structural coherence, is computed using keys $K'$ and values $V'$ derived from $f$ itself.
The {Detect Module} aggregates these maps over the active time window to predict the spatial risk mask $M_t$:
\begin{equation}
    M_t = \text{Norm}(\text{Reshape}(\text{vec}(A_{\text{cross}})^\top \cdot A_{\text{self}}))
\end{equation}
where $\text{vec}(\cdot)$ flattens the map. This fusion allows the module to refine raw semantic activations with precise object boundaries.

The Detect module is trained with direct supervision to ensure accurate localization. We use a combination of Dice loss and L1 loss on a small set of annotated images:
\begin{equation}
     \mathcal{L}_{\text{detect}} = \lambda_{\text{dice}}\mathcal{L}_{\text{Dice}}(A_{\text{cross}}, M_{gt}) + \lambda_{\text{l1}}\mathcal{L}_{\text{L1}}(M_t, M_{gt})
\end{equation}
where $M_t$ is the predicted mask at timestep $t$ and $M_{gt}$ is the ground truth. This supervised pre-training allows for efficient few-shot adaptation to specific visual risks.

\subsection{Preference-Aligned Suppression}
Once the risk mask $M_t$ is determined, the Suppress Module $\mathcal{S}$ intervenes to neutralize harmful semantics.

Instead of relying on rigid replacement words, we align the module with human safety preferences using Direct Preference Optimization (DPO). 
We first use GPT-4 to generate pairs of prompts (unsafe vs. safe). These prompts are fed into the Stable Diffusion model to generate a dataset of preference image pairs: a preferred ``safe'' image $y_w$ and a dispreferred ``unsafe'' image $y_l$.
We train the Suppress Module directly in the latent space. The DPO loss is formulated as:
\begin{equation}
    \mathcal{L}_{\text{DPO}} = -\mathbb{E} \left[ \log\sigma\left(\beta \left( r_\theta(y_w, c) - r_\theta(y_l, c) \right) \right) \right],
    \label{eq:dpo_loss}
\end{equation}
where \(\sigma\) is the sigmoid function, \(\beta\) is a temperature parameter, and \(r_\theta(y, c)\) is the implicit reward function. The reward measures the improvement of our policy model \(\epsilon_\theta\) over the reference model \(\epsilon_{\text{ref}}\) in predicting the noise \(\epsilon\) for a given noisy sample \(z_t\). It is defined as the difference in MSE:
\begin{equation}
    r_\theta(y, c) = \mathbb{E}_{t, \epsilon} \left[ \| \epsilon - \epsilon_{\text{ref}}(z_t, t, c) \|_2^2 - \| \epsilon - \epsilon_\theta(z_t, t, c) \|_2^2 \right].
    \label{eq:dpo_reward}
\end{equation}
Optimizing this objective encourages our model's denoising error to be \textit{smaller} for preferred images (\(y_w\)) and \textit{larger} for rejected images (\(y_l\)), effectively steering the generation trajectory toward safety.

At inference, we perform a hard fusion to strictly preserve the background. Let $V_{orig}$ be the features from the frozen U-Net and $V_{safe}$ be the output from our Suppress Module. The final feature $V_{new}$ is:
\begin{equation}
    V_{new} = V_{orig} \odot (1 - M_t) + V_{safe} \odot M_t
\end{equation}
This guarantees that regions where $M_t \approx 0$ (background) remain mathematically identical to the original output.

\section{Experiments}

We evaluate SafeCtrl to demonstrate its superiority in navigating the critical trade-off between safety, context preservation, and robustness. 

\subsection{Experimental Setup}
\textbf{Datasets and Metrics.} 
We utilize three benchmarks to evaluate safety, fidelity, and robustness. We employ {I2P}~\cite{schramowski2023safe} to assess diverse risks. We report the {inappropriateness ratio} ($R$), defined as the proportion of images flagged by either NudeNet or Q16 classifiers.
On {COCO-30k}~\cite{lin2014microsoft}, we report FID and CLIP scores.
We use {Ring-A-Bell}~\cite{tsai2023ring} to measure the Attack Success Rate (ASR) against adversarial prompts.

To rigorously quantify the trade-off between safety and utility, we introduce the {H-Score} .
We define the {Safety Score} as $S = 1 - R$,  where $R$ is the unsafe ratio.
To measure {Generation Utility} ($U$), we aggregate FID and CLIP. Since these metrics have different scales and polarities, we apply Min-Max normalization based on the range of all compared methods:
\begin{equation}
\small
    F' = 1 - \frac{\text{FID} - \text{FID}_{\min}}{\text{FID}_{\max} - \text{FID}_{\min}}, \quad
    C' = \frac{\text{CLIP} - \text{CLIP}_{\min}}{\text{CLIP}_{\max} - \text{CLIP}_{\min}}
\end{equation}
The utility is the average $U = \frac{1}{2} (F' + C')$. 

The H-Score is the harmonic mean of Safety and Utility:
\begin{equation}
H = \frac{2 \cdot S \cdot U}{S + U}, 
\end{equation}
which penalizes methods that sacrifice one aspect for the other.

\textbf{Baselines.} 
We compare SafeCtrl against state-of-the-art methods across different paradigms: global inference guidance (SLD~\cite{schramowski2023safe}), global fine-tuning (ESD~\cite{gandikota2023erasing}, UCE~\cite{gandikota2024unified}, and AlignGuard~\cite{liu2025alignguard}), and localized control (Concept Replacer, CR~\cite{zhang2025concept}). All experiments are conducted using Stable Diffusion v2.1~\cite{rombach2022high} as the base model.

The training of SafeCtrl is highly efficient. {By leveraging the rich semantic priors of the pre-trained U-Net}, the {Detect Module} achieves robust localization in a few-shot setting ($\sim$10 pixel-annotated images per concept) using a segmentation loss, while the {Suppress Module} employs our Latent DPO strategy on 200 image-level preference pairs. Inference uses the DDIM scheduler with 50 steps.

\begin{figure}[t]
    \centering
    \includegraphics[width=0.95\linewidth]{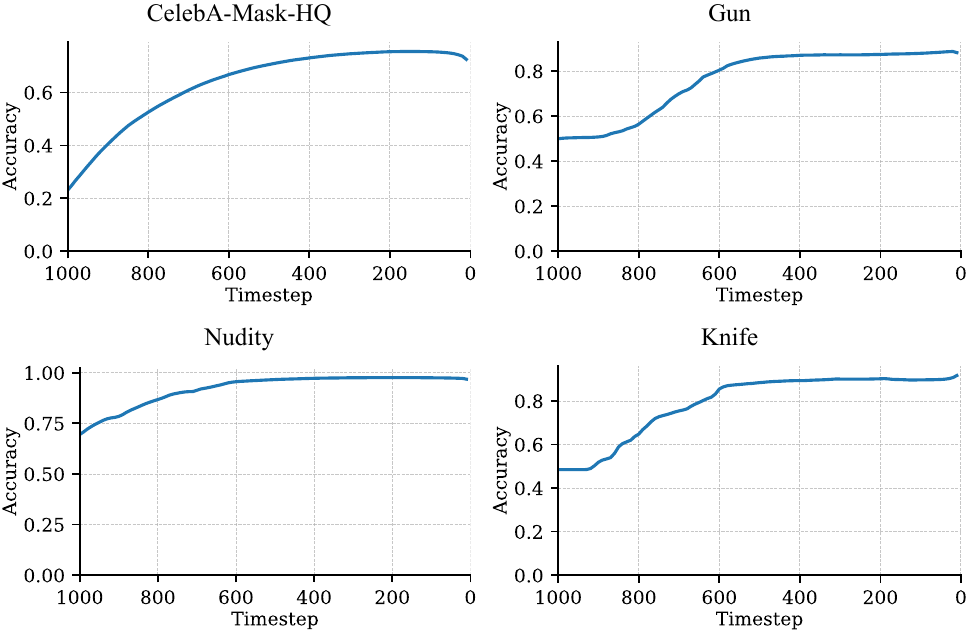} 
    \vspace{-2mm}
    \caption{
        \textbf{Impact of Timesteps on Detection Accuracy.} 
        mIoU scores across concepts consistently improve as the denoising step $t$ decreases (1000 $\to$ 0). 
        This empirically justifies our {Dynamic Scheduling}. We activate detection around $t \in [600, 800]$ where semantic structures stabilize, avoiding computation on early noisy steps.
    }
    \label{fig:ablation_timestep}
    \vspace{-4mm} 
\end{figure}

\begin{figure*}[t!]
    \centering
    \includegraphics[width=0.9\textwidth]{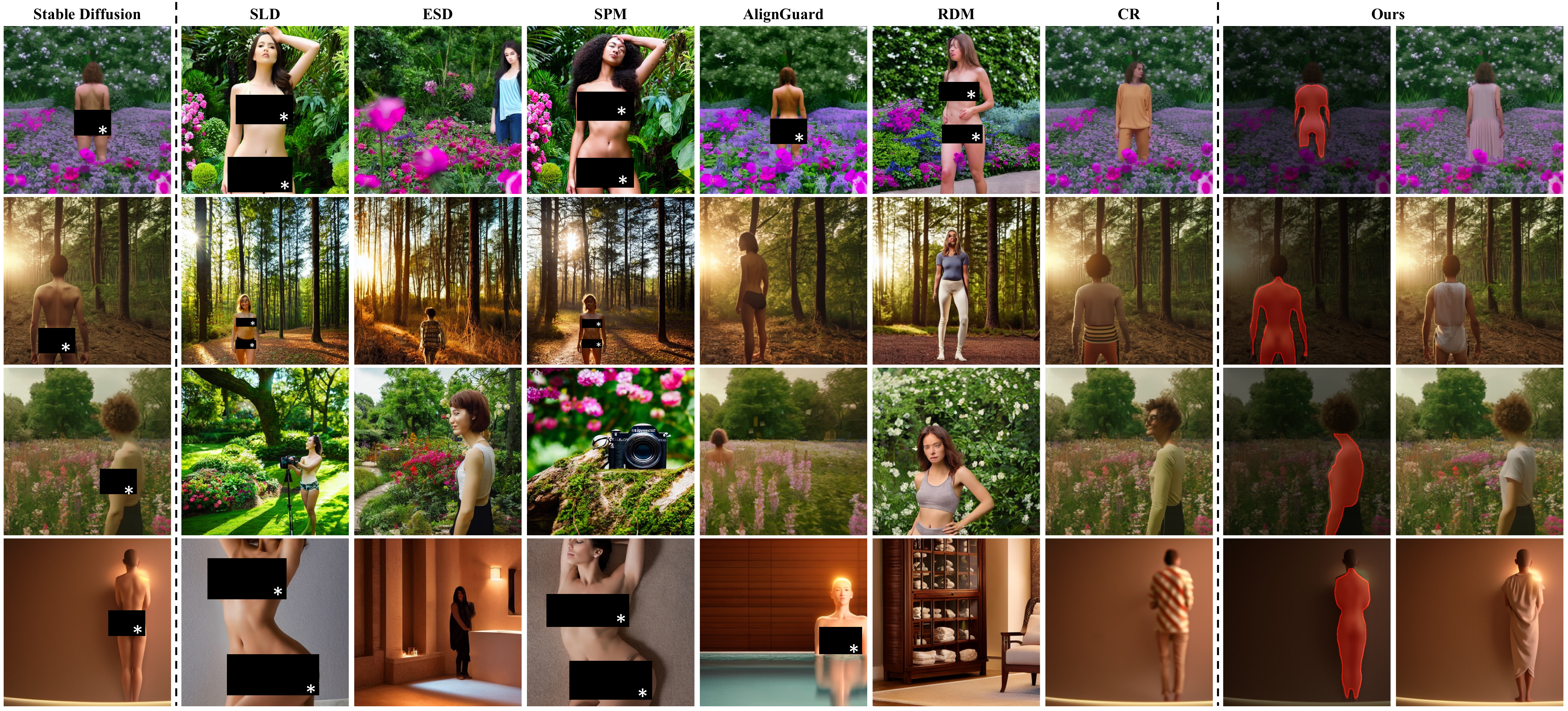}
    \caption{
        \textbf{Qualitative Comparison on Nudity Suppression.} 
        Columns show results using the same seed. 
        SLD often fail to ensure safety, whereas ESD, SPM, AlignGuard and RDM, achieve safety at the cost of altering the original background and identity, leading to context degradation. 
        Although {CR} operates locally, its hard replacement can appear rigid artifacts. 
        {SafeCtrl} combines precise localization (red masks) with natural DPO-based suppression, consistently generating safe outputs while preserving the original artistic intent and background details.
    }
    \label{fig:main_qualitative_comparison}
\end{figure*}

\begin{table}[t]
    \centering
    \caption{
        \textbf{Quantitative Comparison on I2P and COCO-30K.} 
        We report the \textbf{Overall Unsafe Ratio} (lower $\downarrow$ is better), 
        \textbf{FID} (lower $\downarrow$), and \textbf{CLIP} (higher $\uparrow$).
        The \textbf{H-Score} (higher $\uparrow$) quantifies the trade-off, calculated as the harmonic mean of Safety ($1-\text{Overall}$) and normalized Utility.
        \textbf{SafeCtrl} achieves the best safety--utility trade-off.
    }
    \label{tab:main_results}
    \resizebox{\columnwidth}{!}{
    \setlength{\tabcolsep}{5pt}
    \begin{tabular}{l|cccc}
        \toprule
        \textbf{Method} & \textbf{Overall} $\downarrow$ & \textbf{FID} $\downarrow$ & \textbf{CLIP} $\uparrow$ & \textbf{H-Score} $\uparrow$ \\ 
        \midrule
        Original SD & 0.40 & \textbf{14.30} & \textbf{0.2626} & 0.750 \\
        SLD~(CVPR'23) & 0.19 & 18.22 & 0.2543 & 0.640 \\
        ESD~(ICCV'23) & 0.18 & 17.34 & 0.2381 & 0.402 \\
        SPM~(CVPR'24) & 0.34 & 14.77 & 0.2581 & 0.751 \\ 
        AlignGuard~(ICCV'25) & 0.14 & 20.80 & 0.2554 & 0.501 \\
        RDM~(WWW'25) & 0.12 & 15.13 & 0.2588 & 0.869 \\ 
        CR~(CVPR'25) & \underline{0.12} & 15.15 & 0.2551 & 0.828 \\ 
        \midrule
        \textbf{Ours (SafeCtrl)} & \textbf{0.11} & \underline{15.03} & \underline{0.2616} & \textbf{0.906} \\ 
        \bottomrule
    \end{tabular}
    }
\end{table}

\begin{table}[h]
    \centering
    \caption{
        \textbf{Localization Accuracy (mIoU $\uparrow$) on Proxy Benchmarks.}
        Evaluating under a 10-shot setting. SafeCtrl outperforms baselines including the heavyweight CR, verifying precise region awareness.
    }
    \label{tab:localization}
    \footnotesize
    \resizebox{\columnwidth}{!}{
    \setlength{\tabcolsep}{4pt} 
    \begin{tabular}{l|ccc|c} 
    \toprule
    \textbf{Method} & \textbf{Pascal-Car} & \textbf{CelebA-HQ} & \textbf{Pascal-Horse} & \textbf{Avg.} \\
    \midrule
    ReGAN & 52.2 & 69.9 & 54.3 & 58.8 \\
    SLiMe & 68.3 & 75.7 & 63.3 & 69.1 \\
    SegDDPM* & - & 78.0 & 58.7 & - \\
    CR & 69.3 & 78.1 & 64.3 & 70.6 \\
    \midrule
    \textbf{Ours (SafeCtrl)} & \textbf{72.1} & \textbf{78.3} & \textbf{65.7} & \textbf{72.0} \\
    \bottomrule
    \end{tabular}
    }
    \vspace{-10pt} 
\end{table}

\begin{table}[t]
    \centering
    \caption{
        \textbf{Robustness against Adversarial Prompts (Ring-A-Bell).} 
        We report the Unsafe Ratio ($\downarrow$). 
        \textbf{SafeCtrl} maintains robust defense, achieving performance on par with CR.
    }
    \label{tab:robustness}
    \resizebox{\columnwidth}{!}{
    \setlength{\tabcolsep}{3.5pt} 
    \begin{tabular}{l|cccccccc}
        \toprule
        \textbf{Method} & \textbf{SD} & \textbf{SLD} & \textbf{ESD} & \textbf{SPM} & \textbf{AlignG.} & \textbf{RDM} & \textbf{CR} & \textbf{Ours} \\
        \midrule
        \textbf{Unsafe} $\downarrow$ & 0.95 & 0.97 & 0.54 & 0.65 & 0.63 & 0.32 & \textbf{0.27} & \underline{0.28} \\
        \bottomrule
    \end{tabular}
    }
\end{table}

\begin{table}[h]
    \centering
    \caption{
        \textbf{Efficiency Analysis.} 
        Unlike CR which duplicates the U-Net ($\sim$860M params), SafeCtrl is a lightweight plugin ($\sim$75M), offering comparable safety with significantly lower overhead.
    }
    \label{tab:efficiency}
    \footnotesize
    \resizebox{0.9\columnwidth}{!}{
    \setlength{\tabcolsep}{6pt} 
    \begin{tabular}{l|c|c}
    \toprule
    \textbf{Method} & \textbf{Total Added Params} $\downarrow$ & \textbf{Latency (s)} $\downarrow$ \\
    \midrule
    Original SD & - & 10.63 \\
    CR & $\sim$860 M & 12.70 \\
    \textbf{Ours (SafeCtrl)} & \textbf{$\sim$75 M} & \textbf{11.77} \\
    \bottomrule
    \end{tabular}
    }
    \vspace{-10pt}
\end{table}

\begin{figure}[t]
    \centering
    \includegraphics[width=0.9\linewidth]{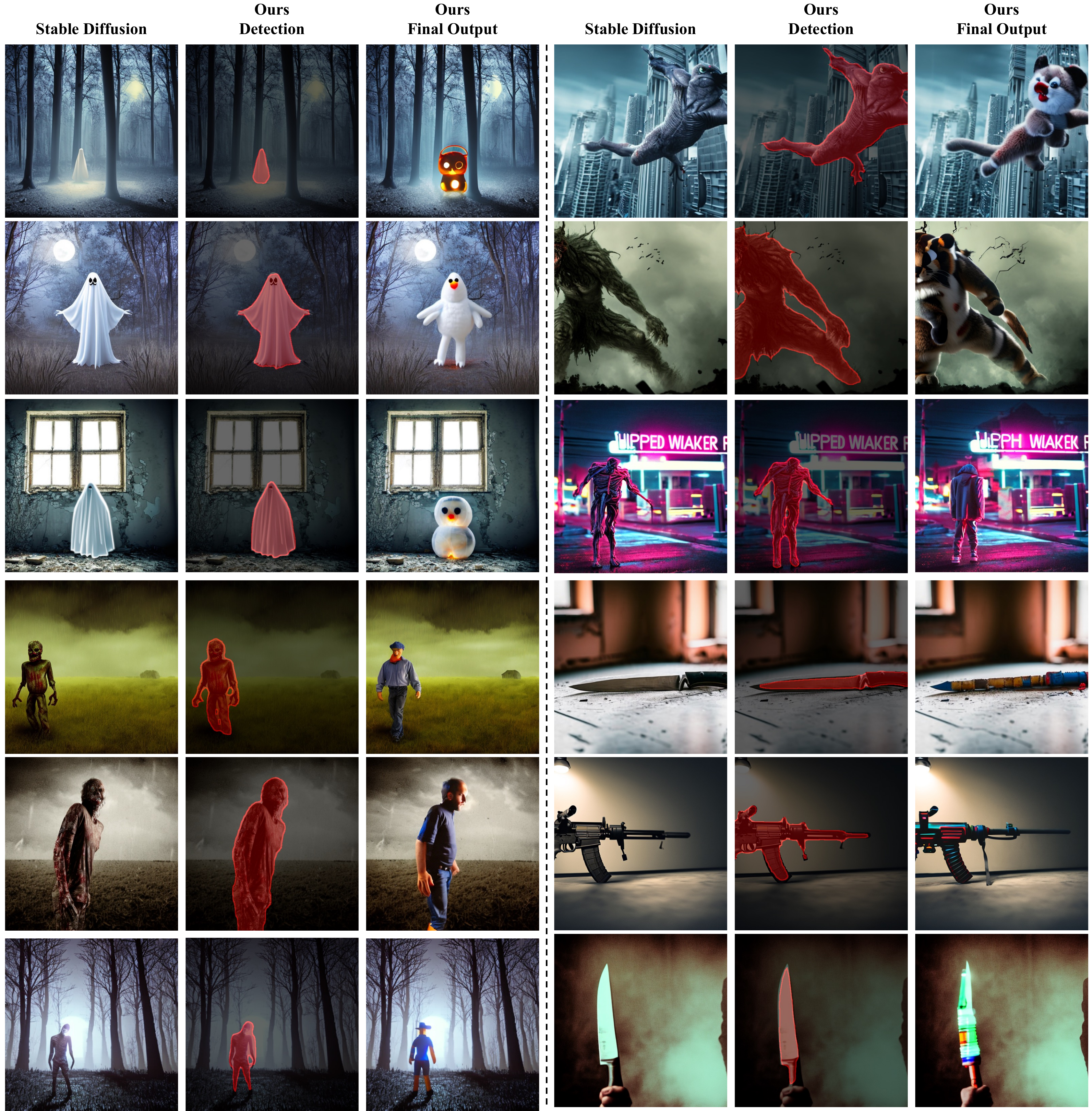}
    \caption{
        {Generalization to Diverse Visual Risks.} 
        SafeCtrl effectively generalizes beyond nudity to mitigate other distinct visual threats. 
        Suppression of abstract horror concepts (e.g., ghosts, zombies, monsters) into benign objects (e.g., dolls, humans) or suppression of concrete violence/weapons (e.g., knives, guns) into safe alternatives.
    }
    \label{fig:generalization}
    \vspace{-4mm} 
\end{figure}

\subsection{Quantitative Results}
Figure~\ref{fig:app_nudenet_grid_2x4} breaks down the unsafe detections by specific NudeNet\cite{bedapudi2019nudenet} categories.
As observed, the original Stable Diffusion (SD) exhibits high risks, particularly in categories like `Breasts (F)' (177 detections) and `Belly' (159).
While global methods like SPM and AlignGuard reduce overall counts, they often fail to completely eradicate specific anatomical features .
In contrast, {SafeCtrl} demonstrates precise suppression capability. For the most distinct risk category (`Breasts (F)'), our method reduces detections to a negligible count of {10}, significantly outperforming other method.
This confirms that our attention-guided module effectively identifies and neutralizes specific visual features.

\textbf{Trade-off Metric (H-Score).} 
Table~\ref{tab:main_results} presents the comparison against state-of-the-art methods.
SafeCtrl achieves the lowest {Overall Unsafe Ratio} of {0.11} on the I2P benchmark, which aggregates seven distinct risk categories. 
It outperforms global fine-tuning methods like ESD (0.18) and surpasses recent SOTA methods like RDM and Concept Replacer.
Unlike global methods that degrade image quality (e.g., AlignGuard's FID of 20.80), SafeCtrl maintains an FID of {15.03} and a CLIP score of {0.2616}, which are closest to the original Stable Diffusion.
Crucially, SafeCtrl achieves the highest {H-Score} ({0.90}), outperforming Concept Replacer (0.83). 
This confirms our method offers the superior safety intervention, effectively balancing strict suppression with context preservation.

\textbf{Localization Accuracy.}
The efficacy of our framework hinges on precise risk identification.
We quantitatively evaluate localization performance on three proxy benchmarks (Pascal-Car, CelebA-HQ, Pascal-Horse) under a strict {10-shot setting}.
We compare SafeCtrl against state-of-the-art few-shot/unsupervised methods (ReGAN~\cite{tritrong2021repurposing}, SLiMe~\cite{khani2023slime}) and {Concept Replacer (CR)}~\cite{zhang2025concept}.
As shown in Table~\ref{tab:localization}, SafeCtrl achieves the highest average mIoU of {72.0}, outperforming SLiMe  and surpassing  CR.
It confirms that by cleverly repurposing the U-Net's internal attention features, we achieve state-of-the-art localization accuracy efficiently, without the need for the duplicated U-Net architecture required by CR.

\textbf{Adversarial Robustness}
We evaluate robustness using the {Ring-A-Bell} benchmark~\cite{tsai2023ring}, which applies adversarial attacks to prompts.
As shown in Table~\ref{tab:robustness}, other methods suffer catastrophic failure. For instance, {AlignGuard} and {ESD} see their unsafe ratios spike to {0.63} and {0.54} respectively, as they rely heavily on the text embeddings. When tokens are obfuscated, the suppression fails to trigger.
In contrast, {SafeCtrl maintains a robust defense (0.28)}, effectively tying with the Concept Replacer (0.27). 
This validates that anchoring detection on internal visual features, which remain activated by the visual manifestation of unsafe concepts even when text is garbled, providing a much more reliable safety method than text-dependent methods.

\textbf{Efficiency Analysis.}
To validate the practicality of our modular approach, we compare efficiency with our main architectural competitor, CR (Table~\ref{tab:efficiency}).
While both methods achieve localized control, CR's reliance on a duplicated U-Net results in a massive memory burden ($\sim$860M added parameters).
In contrast, SafeCtrl reuses internal features, adding only {$\sim$75M parameters}—an order of magnitude smaller. 
Coupled with lower inference latency (11.77s vs. 12.70s), SafeCtrl proves to be a highly deployment-friendly solution without the resource heaviness of prior localized methods.

\textbf{Why Dynamic Scheduling?} 
Our method activates the Detect Module only during the window $[T_{\text{start}}, T_{\text{switch}}]$.
Figure~\ref{fig:ablation_timestep} provides the empirical justification for setting $T_{\text{start}} \approx 700$.
First, regarding {accuracy}, we observe that semantic structures emerge rapidly; detecting before this window ($t > T_{\text{start}}$) yields low mIoU due to noise. The spatial risk mask converges quickly, so a short duration to $T_{\text{switch}}$ is sufficient.
Second, regarding {generative consistency}, early intervention is critical. Detecting and suppressing after this window ($t < T_{\text{switch}}$, e.g., late stages) would act on already-solidified image structures, leading to unnatural artifacts.
By targeting this critical semantic formation window, SafeCtrl achieves the optimal balance of precision and consistency.

\subsection{Qualitative Analysis.}
Figure~\ref{fig:main_qualitative_comparison} provides visual evidence corroborating our quantitative findings. 
As observed, global inference methods SLD often fail to ensure safety. While global fine-tuning methods such as ESD, AlignGuard and RDM successfully remove explicit content, they suffer from severe {context degradation}. For instance, in the second row, ESD completely alters the lighting and tree structure of the forest, explaining its lower fidelity and H-Score. 
In contrast, SafeCtrl demonstrates superior context preservation. The intermediate detection masks (visualized in red) confirm that our module precisely isolates risk regions without spilling over to the environment. This precision allows our DPO-trained Suppress Module to perform a natural transformation that blends seamlessly with the frozen context, avoiding the rigid artifacts sometimes seen in CR. 

\textbf{Generalization To Other Concepts.}
A key advantage of our region-aware framework is its adaptability. SafeCtrl leverages semantic attention features that generalize to diverse visual objects.
As shown in Figure~\ref{fig:generalization}, we demonstrate efficacy across two distinct categories: {Horror/Supernatural} (Left) and {Violence/Weapons} (Right).
Whether dealing with the organic, irregular shapes of {Zombies} and {Monsters}, or the rigid, geometric structures of {Guns} and {Knives}, our Detect Module generates precise spatial masks (visualized in red).
Consequently, the Suppress Module transforms these unsafe entities into safe alternatives (e.g., turning a gun into a toy, or a ghost into a doll) without altering the surrounding environment. This confirms that SafeCtrl is a versatile solution for broad-spectrum visual safety.

\section{Conclusion}
In this work, we introduced SafeCtrl, a novel framework designed to resolve the fundamental safety-capability trade-off in text-to-image diffusion. 
We proposed a flexible, {region-aware} `Detect-Then-Suppress' paradigm.
The strength of this paradigm lies in its attention-guided Detect module, which acts as a precise risk assessor capable of identifying diverse {localizable visual concepts} .
Furthermore, we demonstrated that our Suppress module, trained via image-level Direct Preference Optimization, effectively neutralizes risks without requiring pixel-level annotations.
Extensive experiments validate that SafeCtrl significantly outperforms state-of-the-art methods in achieving a superior trade-off between safety and {context preservation}. 

\section{ACKNOWLEDGMENT}
This work was funded in part by the the Key Research and Development Programme of Ningbo’s “Science and Technology Innovation Yongjiang 2035” Plan under Grant 2025Z054 and the Jiangsu Provincial Natural Science Foundation of China under Grant BK20240291.

\bibliographystyle{IEEEbib}
\bibliography{icme2026references}

\end{document}